\documentclass{llncs}
\usepackage{graphicx}
\usepackage{mathtools}

\usepackage{epsfig}
\usepackage{amsmath}
\usepackage{amssymb}
\usepackage{multirow}
\usepackage{color}
\usepackage{array}
\usepackage{subfig}
\usepackage[ruled,vlined,linesnumbered]{algorithm2e}
\usepackage{soul}

\newcolumntype{L}[1]{>{\raggedright\let\newline\\\arraybackslash\hspace{0pt}}m{#1}}
\newcolumntype{C}[1]{>{\centering\let\newline\\\arraybackslash\hspace{0pt}}m{#1}}
\newcolumntype{R}[1]{>{\raggedleft\let\newline\\\arraybackslash\hspace{0pt}}m{#1}}

\newcommand*{\affmark}[1][*]{\textsuperscript{#1}}

\begin{document}

\title{Adversarial Domain Adaptation for Classification of Prostate Histopathology Whole-Slide Images}

\author{Jian Ren\affmark[1], Ilker Hacihaliloglu\affmark[2], Eric A. Singer\affmark[3], David J. Foran\affmark[3], Xin Qi\affmark[3]}

\institute{\affmark[1]Department of Electrical and Computer Engineering, Rutgers University, USA\\
\affmark[2]Department of Biomedical Engineering, Rutgers University, USA\\
\affmark[3]Rutgers Cancer Institute of New Jersey, USA
}

\maketitle             

\begin{abstract}
Automatic and accurate Gleason grading of histopathology tissue slides is crucial for prostate cancer diagnosis, treatment, and prognosis. Usually, histopathology tissue slides from different institutions show heterogeneous appearances because of different tissue preparation and staining procedures, thus the predictable model learned from one domain may not be applicable to a new domain directly. Here we propose to adopt unsupervised domain adaptation to transfer the discriminative knowledge obtained from the source domain to the target domain without requiring labeling of images at the target domain. The adaptation is achieved through adversarial training to find an invariant feature space along with the proposed Siamese architecture on the target domain to add a regularization that is appropriate for the whole-slide images. We validate the method on two prostate cancer datasets and obtain significant classification improvement of Gleason scores as compared with the baseline models.

\end{abstract}
\section{Introduction}
Prostate cancer is the most common non-cutaneous malignancy and affects 1
in 7 men in the United States~\cite{ferlay2015cancer}. Gleason scores, graded from whole-slide images (WSIs), have been shown to serve as one of the best predictors for prostate
cancer diagnosis~\cite{epstein2016contemporary}. Gleason grading is crucial for studying disease onset, progression and decision making for targeted therapy. However, Gleason grading
is a time-consuming process due to the giga-pixel size of the WSIs. Furthermore, inter- and intra-observer variability errors often arise when pathologists
make diagnosis based on WSIs. In order to provide an objective and quantitative
Gleason grading score, computational methods have been applied for detection,
extraction, and recognition of histopathological patterns. Methods based on convolutional neural networks (CNN) are considered state-of-the-art due to their
high classication rates~\cite{hou2016patch}\cite{litjens2016deep}\cite{otalora2015combining}. Most of these studies focus on the supervised
classification. Histopathology WSIs obtained from different institutions usually
present distinct glandular region distributions due to differences in appearance
that may be caused by using different microscope scanners and staining procedures. 
These differences may render the supervised classification model used for predicting the Gleason score for one annotated dataset (source domain) ineffective on another prostate dataset (target domain). A widely used approach to address the challenge is to label new images on the target domain and fine-tune the model trained on source domain~\cite{schmidhuber2015deep}. 
Instead, methods that can learn from existing datasets and adapt to new target domains, without the need for additional labeling, are highly desirable. 

Thus in this work, we aim to classify the newly given prostate datasets into low and high Gleason grade through unsupervised learning. To achieve this goal, we adopt the unsupervised domain adaptation paradigm to align the image distributions along the annotated source domain and the unlabeled target domain, where the two domains have the same number of high-level classes~\cite{tzeng2017adversarial}\cite{ganin2016domain}. 
We apply adversarial training to minimize the distribution discrepancy at the feature space between the domains, with the loss function adopted from the Generative Adversarial Network (GAN)~\cite{goodfellow2014generative}. Furthermore, we developed a Siamese architecture for the target network to serve as a regularization of patches within the WSIs. The proposed method is validated on public prostate datasets and a newly collected local dataset. The experimental results show the approach significantly improves the classification accuracy of Gleason score as compared with the baseline model. To the best of our knowledge, this is the  first  study of  domain adaptation for unsupervised prostate histopathology WSIs classification.

%--------------------------------Figure: network--------------------------------------------
\begin{figure*}[h]
\begin{center}
\includegraphics[width=0.95\columnwidth]{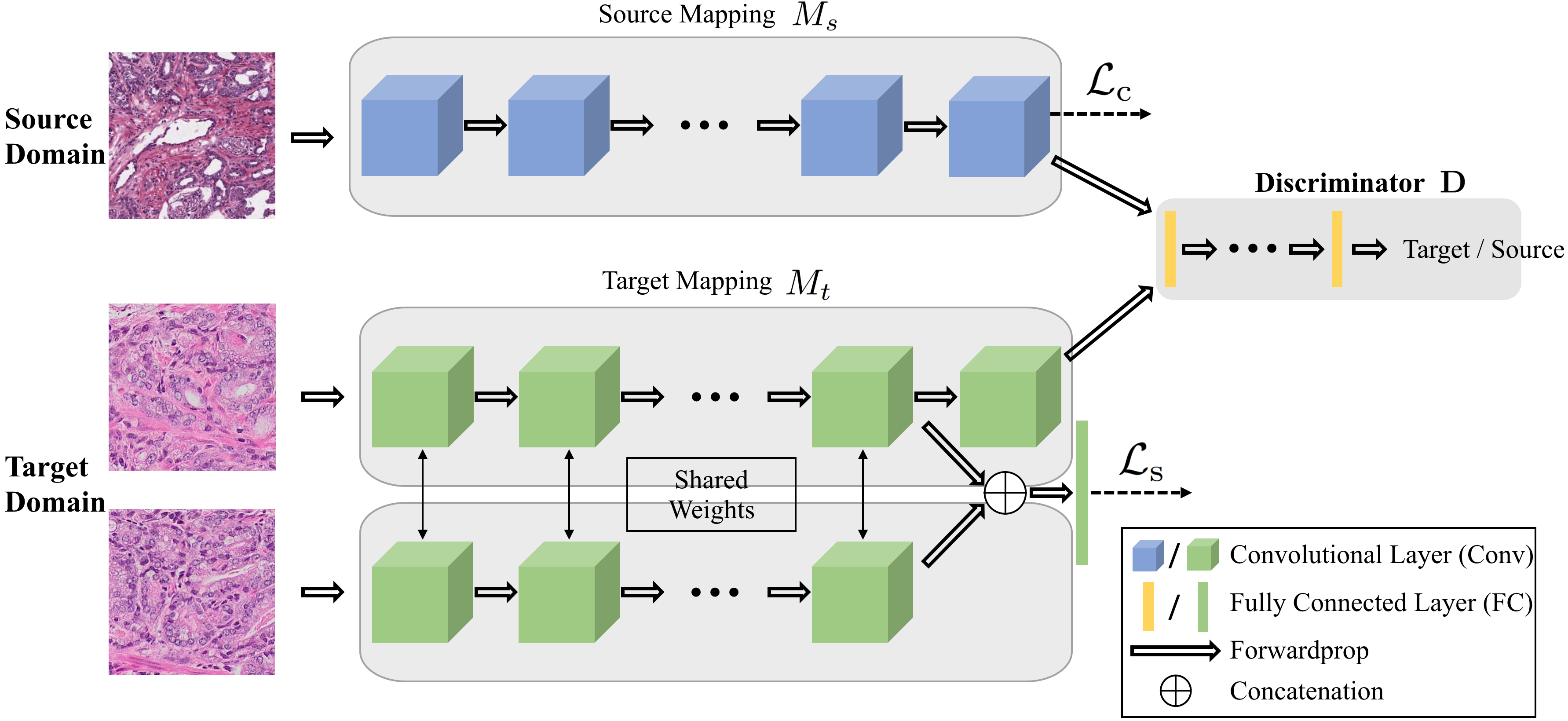}
\caption{The architecture of the networks for the unsupervised domain adaptation. The source network and the target network map the input samples into the feature space. The adaptation is accomplished by jointly training the discriminator and target network using the GAN loss to find the domain invariant feature. A Siamese network at target domain adds constrains for the WSIs.}
\label{network}
\end{center}
 \vspace*{-10mm}
\end{figure*}
%--------------------------------Figure: network--------------------------------------------

\section{Method}
In this section, we present our approach on the unsupervised domain adaptation for the classification of prostate histopathology WSIs, as illustrated in Figure~\ref{network} above.
 \vspace*{-3mm}
\paragraph{\textbf{Problem formulation:}} Formally, we have a source domain distribution  $\mathcal{S}$ that includes $N_s$ labeled prostate histopathology images $\left \{ {(\mathbf{x}_{i}^s, \mathbf{y}_{i}^s)} \right \}_{i=1}^{N_s}$ where $\mathbf{y}_{i}^s$ is one-hot vector denoting the Gleason score, and a target domain distribution $\mathcal{T}$ contains $N_t$ unlabeled prostate histopathology images $\left \{ {(\mathbf{x}_{i}^t)} \right \}_{i=1}^{N_t}$. We use the source domain to generate a feature space through the mapping function $M_s$, and seek to find the mapping $M_t$ at the target domain to obtain a similar feature space with the one from source domain. Thus the Gleason score prediction for the target domain is easily achieved by using the $M_t$.
 \vspace*{-3mm}
\paragraph{\textbf{Learning at source domain:}}
Since the Gleason scores for the prostate images from the source domain are available, 
we train the network on the source domain to get the discriminative feature space using the supervised learning. 
In order to feed the WSIs into the network, we crop them into patches and adopt the cross-entropy loss  $\mathcal{L}_\text{c}$ to optimize the classifier $\mathbf{C}$, with weights as $\theta^S$, to classify the images into low-grade (score as 6 and 7) and high-grade (score higher than 7) Gleason scores, which
are highly related to clinical outcomes.
\begin{equation}
\small
\mathcal{L}_\text{c} =  \mathbb{E}_{\mathbf{x}_s\sim \mathcal{S}}  -\sum_{i=1}^{N_s}\mathbf{y}^s_i \cdot  \hspace{0.1cm} \text{log}\mathbf{C}(M_s(\mathbf{x}_s; \theta^S))
\label{task_loss}
\end{equation}
The majority vote is applied on the cropped patches within each WSI to obtain the final Gleason score for the WSIs.
 \vspace*{-3mm}
\paragraph{\textbf{Adversarial adaptation for target domain:}}
Due to lack of annotations for the training set on the target domain, the $\mathcal{L}_\text{c}$ is only applied on the source domain. To optimize the target network, we leverage the adversarial training to minimize the discrepancy between the feature space of the target domain and the one of the source domain. 
We perform an asymmetric adaptation where the network at the target domain is fine-tuned from the network of the source domain.
Through optimization, the feature space of the target domain learns to mimic the distribution of the source feature space.
Thus the target network is trained to extract the domain invariant features from input samples, which has the same distribution as the source domain.

Adversarial training is achieved by utilizing a GAN loss~\cite{goodfellow2014generative}. Two feature spaces generated from the source network and target network are fed into the discriminator $\mathbf{D}$. $\mathbf{D}$ is trained to map the input feature spaces into a binary domain label, where the true denotes the source domain and false denotes the target domain. Additionally, the target mapping $M_t$, is learned in an adversarial manner to purposely mislead the discriminator by reversing the domain label so that it cannot distinguish between the two feature spaces. Since the mapping parameterization of source model is determined before the adversarial training, we only optimize the target mapping. 
By using adversarial learning, we minimize the discrepancy between the two spaces. Therefore, estimating the Gleason scores for the images from target domain can be implemented by $M_t$.
More specifically, the adversarial loss $\mathcal{L}_{\text{adv}_\mathbf{D}}$ for optimizing the discriminator and the mapping loss $\mathcal{L}_{\text{adv}_M}$ for optimizing the target mapping are represented as:
\begin{equation}
\small
\begin{split}
\underset{\mathbf{D}}{\text{min}} \hspace{0.1cm} \mathcal{L}_{\text{adv}_\mathbf{D}} = - \mathbb{E}_{\mathbf{x}_s\sim \mathcal{S}} \hspace{0.1cm}\text{log}\mathbf{D}(M_s(\mathbf{x}_s; \theta^S); \theta^D)   -\mathbb{E}_{\mathbf{x}_t\sim \mathcal{T}} \hspace{0.1cm}\text{log}(1-\mathbf{D}(M_t(\mathbf{x}_t; \theta^T); \theta^D)
\end{split}
\end{equation}

\begin{equation}
\small
\underset{M_t}{\text{min}} \hspace{0.1cm} \mathcal{L}_{\text{adv}_M} =  - \mathbb{E}_{\mathbf{x}_t\sim \mathcal{T}} \hspace{0.1cm} \text{log}(\mathbf{D}(M_t(\mathbf{x}_t;\theta^T); \theta^D))
\end{equation}

For the adversarial training, we optimize the $\mathcal{L}_{\text{a}}$, where $\mathcal{L}_{\text{a}} = \mathcal{L}_{\text{adv}_\mathbf{D}} + \mathcal{L}_{\text{adv}_M}$.

% %***********************algorithm*****************************************
 \vspace*{-5mm}

\begin{algorithm}
\small
    \SetKwInOut{Input}{Input}
   \Input{Initialized target network from source network with weights $\theta^T$ = $\theta^S$}
    \For{number of training iterations}{
    sample two same number of mini-batches  $\mathbf{x_s}\sim \mathcal{S}$, $\mathbf{x_t}\sim \mathcal{T}$;\\
    obtain the estimation $\mathbf{y} = M_s(\mathbf{x}_s; \theta^S)$,  $\mathbf{y'} = M_t(\mathbf{x}_t; \theta^T)$;\\
    $\theta^D\leftarrow $ back propagate with stochastic gradient $\triangledown\mathcal{L}_{\text{adv}_\mathbf{D}}(\mathbf{y}, \mathbf{y'})$;\\
    $\theta^T\leftarrow $ back propagate with stochastic gradient $\triangledown\mathcal{L}_{\text{adv}_\mathbf{M}}(\mathbf{y'})$;\\
     sample mini-batches with paired of images $\mathbf{x_t^1, x_t^2}\sim \mathcal{T}$;\\
     obtain the estimation $\mathbf{y_f} = f(\mathbf{x_t^1, x_t^2}; \theta^F)$;\\
      $\theta^F\leftarrow $ back propagate with stochastic gradient $\triangledown\mathcal{L}_{\text{s}}(\mathbf{y_f})$;\\
    }
    \caption{Learning Algorithm for the Network at Target Domain}
\label{alg}
\end{algorithm}
 \vspace*{-5mm}

%***************************************************************
 \vspace*{-5mm}
\paragraph{\textbf{Siamese architecture at target domain:}}
Although there are no annotations for the prostate WSIs at the target domain, the cropped patches from the same WSI should still be predicted with the same Gleason score by the target network. While the adversarial loss forces the distribution across two domains to be similar, it can not constrain the target network to determine the similarity of the input patches. Therefore, we introduce a Siamese architecture at target domain to explicitly regularize patches from the same WSI to have the same Gleason score. As shown in Figure~\ref{network}, the two identical networks share the same weights with the input as a pair of images ($\mathbf{x}_t^1$, $\mathbf{x}_t^2$) $\subseteq \mathcal{T} \times \mathcal{T}$. 
The feature maps obtained from the second to the last layer of the two networks are concatenated to serve as the input for a one-layer perceptron to classify the features. Therefore, the input samples are classified by the function $f(\mathbf{x}_t^1, \mathbf{x}_t^2;\theta^F)$, that $f: \mathcal{T} \times \mathcal{T} \mapsto {0, 1}$, where 1 indicates input patches belong to the same WSI while 0 denotes not. We learn the binary classifier $f$ using cross-entropy loss $\mathcal{L}_{\text{s}}$. 

To learn the network at target domain, we adopt a two-stage training process. For the first stage, we train the network at source domain.  For the second stage, we optimize the Siamese network at target domain by applying $\mathcal{L}_{\text{t}}$ where $\mathcal{L}_{\text{t}} = \mathcal{L}_{\text{a}} + \mathcal{L}_{\text{s}}$. The learning algorithm for the target network is shown in Algorithm~\ref{alg}.

\section{Experimental Validation and Results}
Validation of the proposed method is performed in two datasets: (1) publicly available The Cancer Genome Atlas (TCGA) dataset~\cite{kandoth2013mutational}, and (2) a local data set collected from Cancer Institute of New Jersey (CINJ)
after obtaining the institutional review board (IRB) approval.
%---------------------------tcga dataset------------------------------
\begin{table}
\small
\begin{center}
\centerline{
\begin{tabular}{|  C{1.5cm} | C{1.9cm} | C{2cm} |C{2cm} |C{2cm} |C{1.6cm} | }
\hline
 & Gleason 6 & Gleason 7  & Gleason 8 & Gleason 9 & Gleason 10  \\ \hline\hline
\# WSIs& 115 (32)&395 (95)& 94 (20)&128 (24)&4 (0)\\ \hline
\# Patches &  16293 (6517) & 67162 (26583) & 16204 (4968) & 23978 (9606) & 342 (0)\\ \hline
\end{tabular}
}
\end{center}
\caption{The number of WSIs and patches for the prostate histopathology  images from TCGA under different Gleason scores. The images from University of Pittsburgh (UP) are shown in parentheses. }
\label{tcga_number}
 \vspace*{-3mm}
\end{table}
%----------------------------tcga dataset-------------------------------
 \vspace*{-3mm}
\paragraph{\textbf{Dataset}}
In the first unsupervised domain adaptation experiment, we only use the TCGA dataset. The TCGA prostate cancer data includes histopathology WSIs uploaded from 32 institutions that have been acquired at 40$\times$ and 20$\times$ magnifications. We crop the WSIs into patches by the size of 2048$\times$2048. 
We calculate the tissue area on the grayscale images and remove the images with tissue area less than the half of the patch size.
The dataset includes the Gleason scores annotated by pathologists ranging from 6 to 10. As the University of Pittsburgh (UP) has contributed more images than other institutions, we treat the UP as the target domain where the annotations are withheld and the images from other institutions as the source domain, which we denote it as TCGA (w/o UP). We show the total number of WSIs and the cropped patches from TCGA in Table~\ref{tcga_number} and UP in the parentheses. We denote the adaptation as TCGA (w/o UP)  $\to$ UP.
For the second unsupervised domain adaptation experiment, we use all the images from TCGA as the source domain, and the images from CINJ as the target domain. The images from CINJ are acquired at 20$\times$ magnification. More details of the CINJ dataset is shown in Table~\ref{prostate_cinj}. The dataset is labeled by one pathologist with the Gleason scores as 6 or 8. We denote the adaptation as TCGA $\to$ CINJ.
%--------------------------------number of images and baseline------------------------------------------
\begin{table}[!tb]
\small
%     \caption{Global caption}
    \begin{minipage}{.45\linewidth}
      
      	\centering
        \begin{tabular}{|  C{1.5cm} | C{1.6cm} | C{1.6cm} | }
		\hline & Gleason 6 & Gleason 8  \\ \hline\hline
		\# WSIs& 57 &26\\ \hline
		\# Patches &  3933 & 666\\ \hline
		\end{tabular}
        \caption{The number of WSIs and patches for the dataset from CINJ under different Gleason grades.}
        \label{prostate_cinj}
    \end{minipage}%
    \hfill
    \begin{minipage}{.45\linewidth}
      \centering
        \begin{tabular}{|  C{3cm} | C{2.2cm} |  }
		\hline & Accuracy (\%)  \\ \hline\hline
		Previous Study~\cite{jimenez2017convolutional}& 73.5\\ \hline
		TCGA (w/o UP) &  \textbf{76.9} \\ \hline
		TCGA &  \textbf{83.0} \\ \hline
		\end{tabular}
        \caption{The network performance at the source domain. The two source networks both have better performance than~\cite{jimenez2017convolutional}.}
        \label{source_domain}
    \end{minipage} 
 \vspace*{-6mm}
\end{table}
%--------------------------------number of images and baseline------------------------------------------
\vspace*{-3.5mm}
\paragraph{\textbf{Implementation Details}}
For the two sets of experiments, we aim to optimize the network at target domain that could classify the WSIs into low and high Gleason scores.
Thus we divide the TCGA dataset into low Gleason grade for the WSIs with score as 6 and 7, and high Gleason grade for the WSIs with score as 8, 9 and 10.
For the CINJ dataset, the WSIs with Gleason score of 6 belong to the low Gleason grade and Gleason score of 8 belong to high Gleason grade.
The training process is composed of two steps. We first train the binary classification network using the data from the source domain. We use a modified fully convolutional AlexNet~\cite{krizhevsky2012imagenet}, which only contains convolutional layers, as the network for the classification task. All the convolutional layers are followed by the Batch Normalization layer except the last one that gives the prediction. The data from source domain is randomly divided into the training and the testing sets at a ratio of 80\% (validation set is selected from the training set) / 20\%. The patients with more than one WSIs can only contribute the images to the training set or the testing set.
During the training process, the images are resized as 256$\times$256 and randomly cropped to 224$\times$224 to feed into the network. And we train the network from scratch.
The second step is to optimize the Siamese network at target domain. During the second step, we fix the parameters of the source network, and train the target network and the discriminator network at the same time. The feature vectors from the two domains are sent into the discriminator network that contains three fully connected layers. 
And the last layer gives the domain label estimation for the input feature samples. 
The prostate images at the target domain are randomly divided into the training and the testing sets at a ratio of 80\% / 20\%. 

\begin{figure}%
    \centering
    \subfloat[The confusion matrix for UP before adaptation.\label{confusion:UP_b}]{{\includegraphics[width=.22\linewidth]{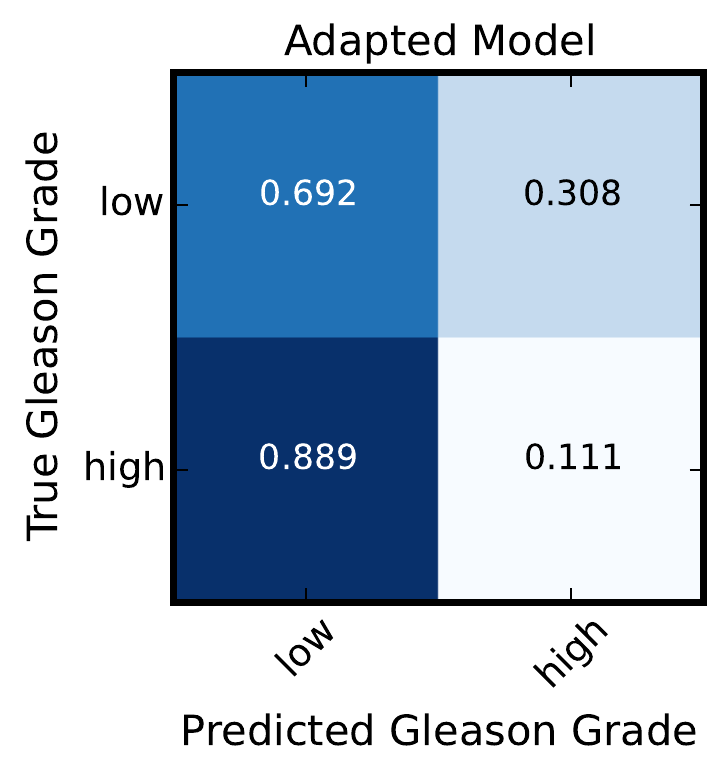} }}%
    \hfill
    \subfloat[The confusion matrix for UP after adaptation.\label{confusion:UP_a}]{{\includegraphics[width=.22\linewidth]{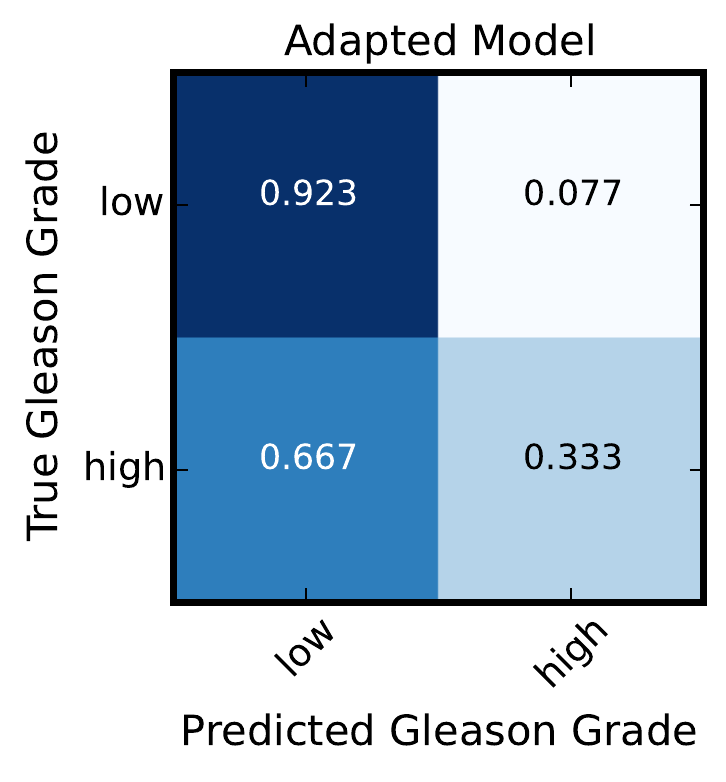} }}%
    \hfill
    \subfloat[The confusion matrix for CINJ before adaptation.\label{confusion:CINJ_b}]{{\includegraphics[width=.22\linewidth]{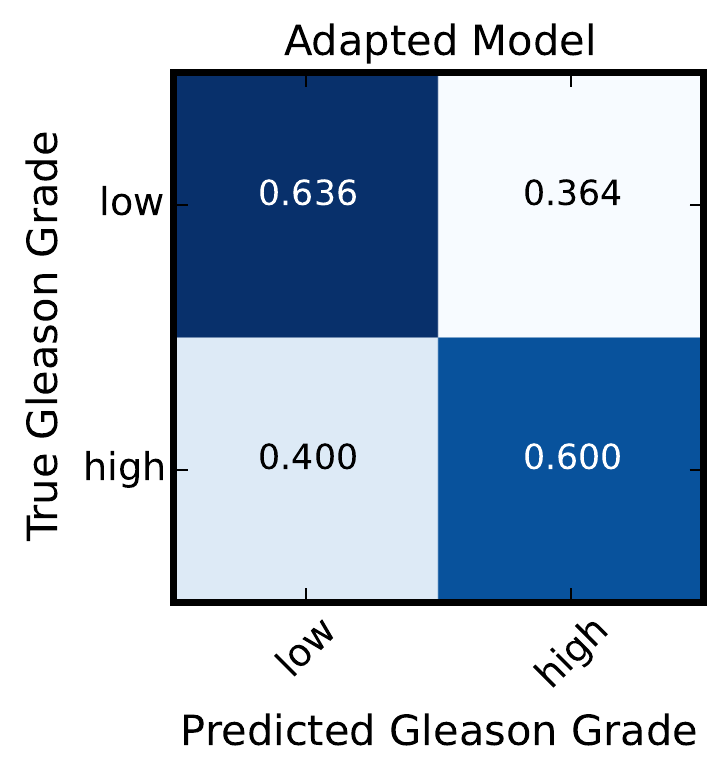} }}%
    \hfill
    \subfloat[The confusion matrix for CINJ after adaptation.\label{confusion:CINJ_a}]{{\includegraphics[width=.22\linewidth]{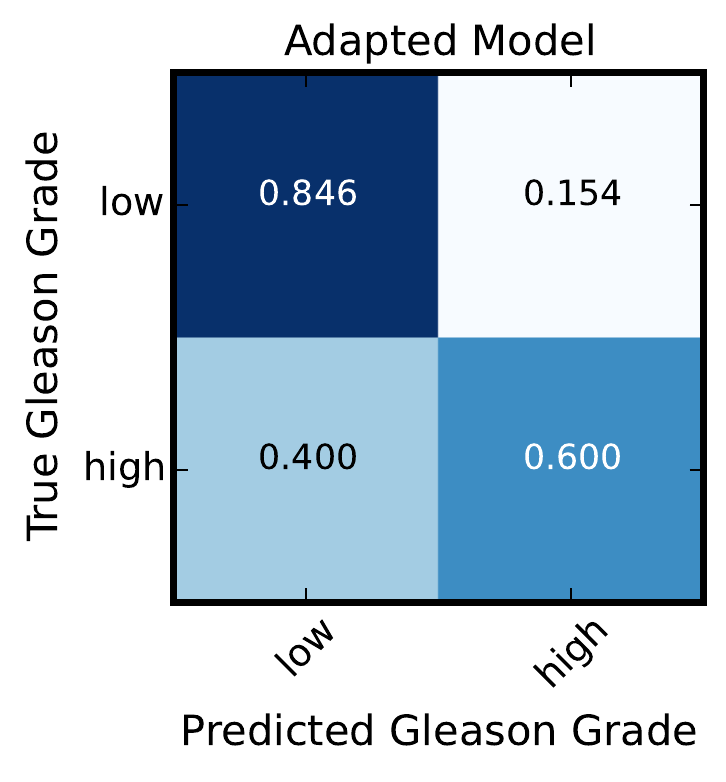} }}%
    \caption{The confusion matrix of the target network before and after the adaptation for  TCGA (w/o UP) $\to$ UP and  TCGA $\to$ CINJ .}%
    \label{fig:example}%
     \vspace*{-8mm}
\end{figure}

 \vspace*{-3mm}
\paragraph{\textbf{Source network performance}}
As the training process contains two steps, we first show the performance of the network at the source domain. The comparison between the source network and the previous study~\cite{jimenez2017convolutional} is shown in Table~\ref{source_domain}. From the results, we can see both of our models have better performance than~\cite{jimenez2017convolutional}. However, the study at~\cite{jimenez2017convolutional} uses less WSIs than ours and the network with the best performance reported in~\cite{jimenez2017convolutional} is wider and deeper than our study.
Although such differences lead to biased comparison, it still demonstrates the source domain network is well trained to classify the TCGA prostate images into low Gleason score and high Gleason score.
 \vspace*{-3mm}
\paragraph{\textbf{Adaptation of TCGA (w/o UP) $\to$ UP}}
In order to prove the effectiveness of the knowledge transfer from source domain to the target domain, we show the quantitative results for TCGA (w/o UP) $\to$ UP in Table~\ref{tcga_up}. We can see that  due to the different image distribution for the TCGA (w/o UP) and UP, the network learned from  TCGA (w/o UP) is not working appropriately on UP. But through the unsupervised adaptation, we could effectively adapt the discriminative knowledge from TCGA (w/o UP) to the UP without requiring additional annotations. We further calculate the statistically significance of the accuracy improvement between the adapted network and the baseline network using McNemar Test~\cite{fagerland2013mcnemar} and demonstrates the improvement of classification accuracy is statistically significant with a p-value as 0.039.
In addition, we show the result of the ablation study in Table~\ref{tcga_up} that using $\mathcal{L}_{\text{t}}$ achieves better classification accuracy than $\mathcal{L}_{\text{a}}$ only.  The confusion matrices for the adaptation are shown in Figure~\ref{confusion:UP_b}-\ref{confusion:UP_a}. After the adaptation, the classification accuracy for both WSIs of low and high Gleason scores are significantly improved.

\begin{figure}%
    \centering
    \subfloat[Example image from CINJ with Gleason grade 8.\label{quan:o}]{{\includegraphics[width=.2\linewidth]{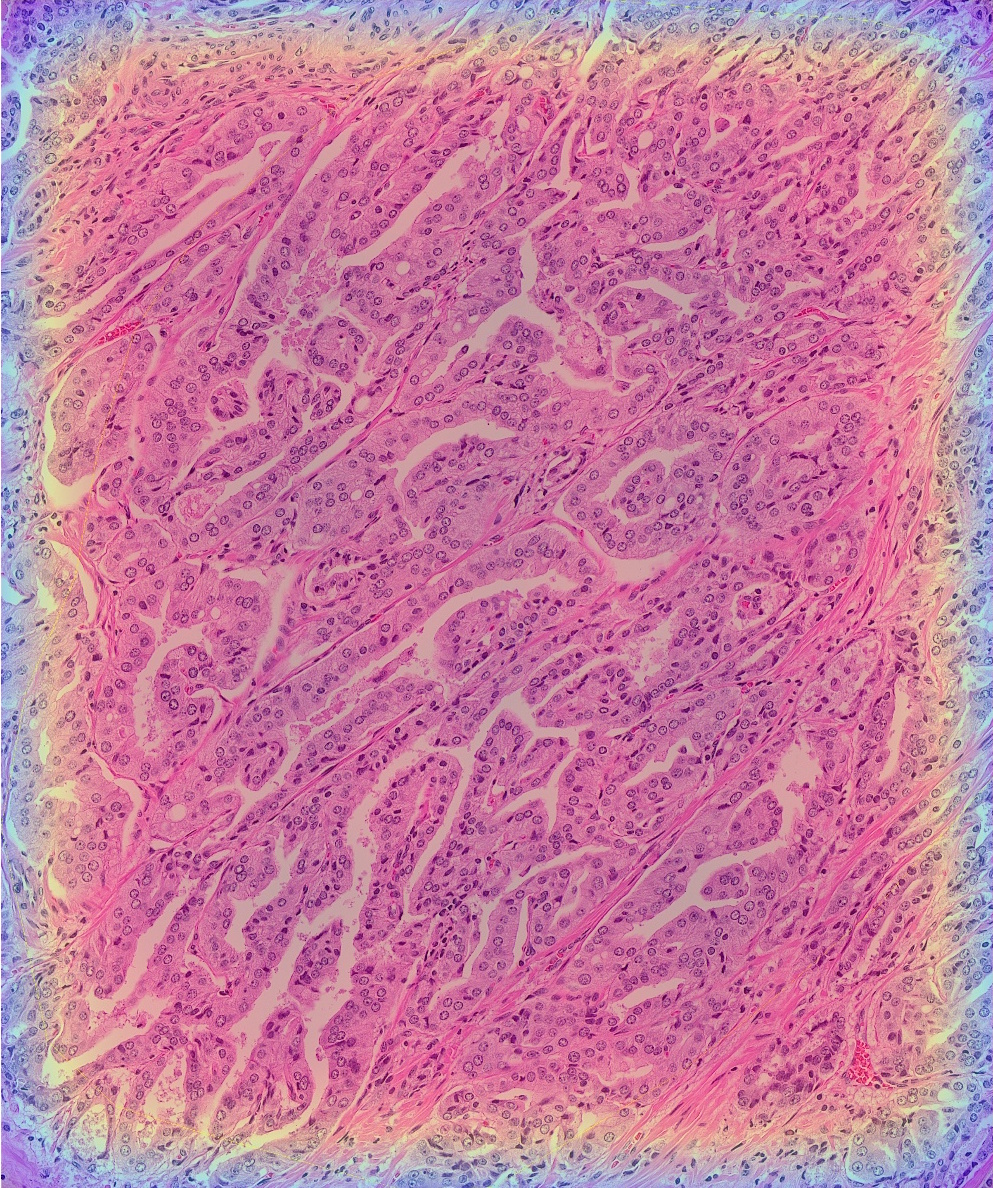} }}%
    \hfill
    \subfloat[Heatmap generated from the baseline model.\label{quan:b}]{{\includegraphics[width=.2\linewidth]{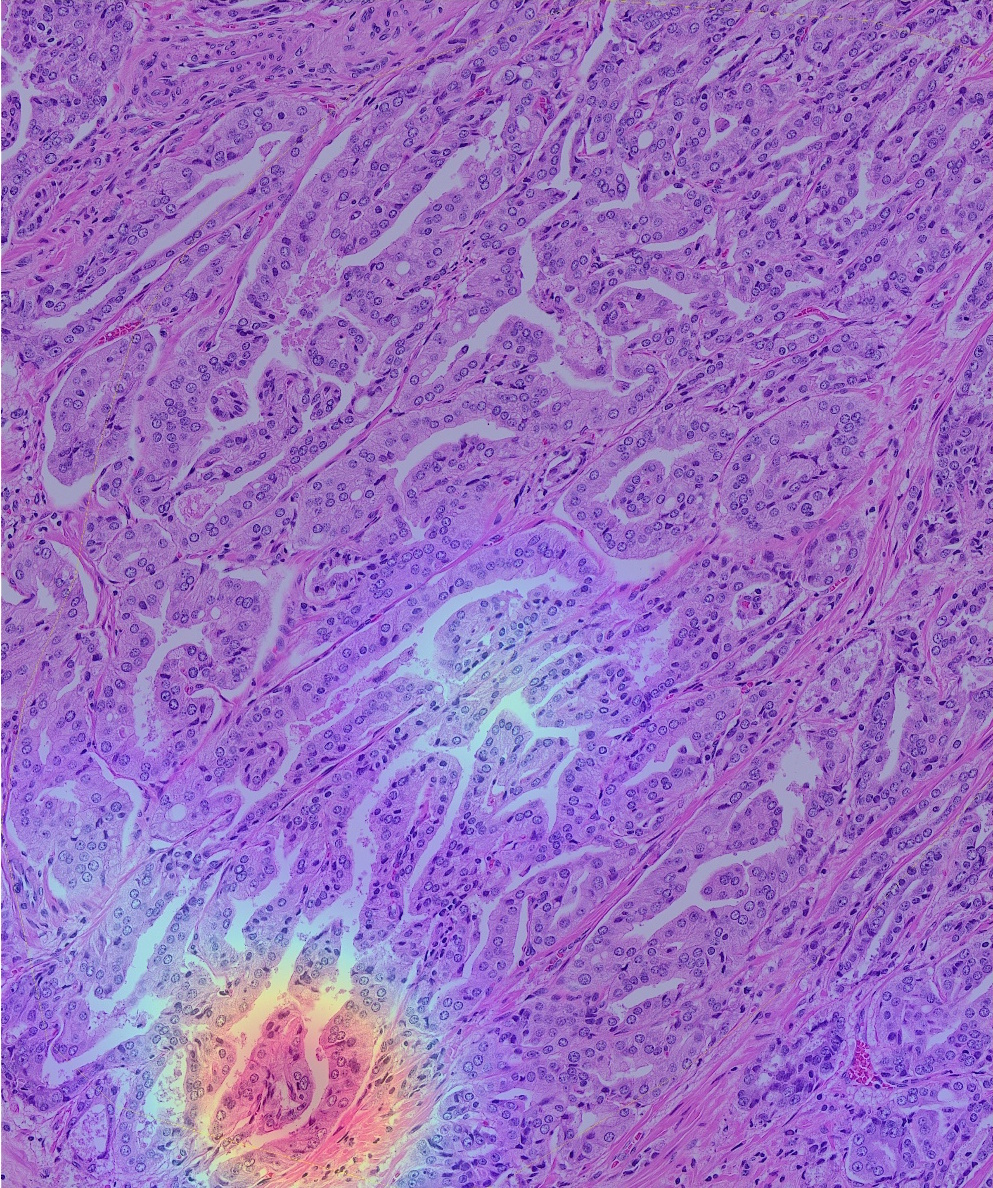} }}%
    \hfill
    \subfloat[Heatmap generated from the model optimized with $\mathcal{L}_{\text{a}}$.\label{quan:a}]{{\includegraphics[width=.2\linewidth]{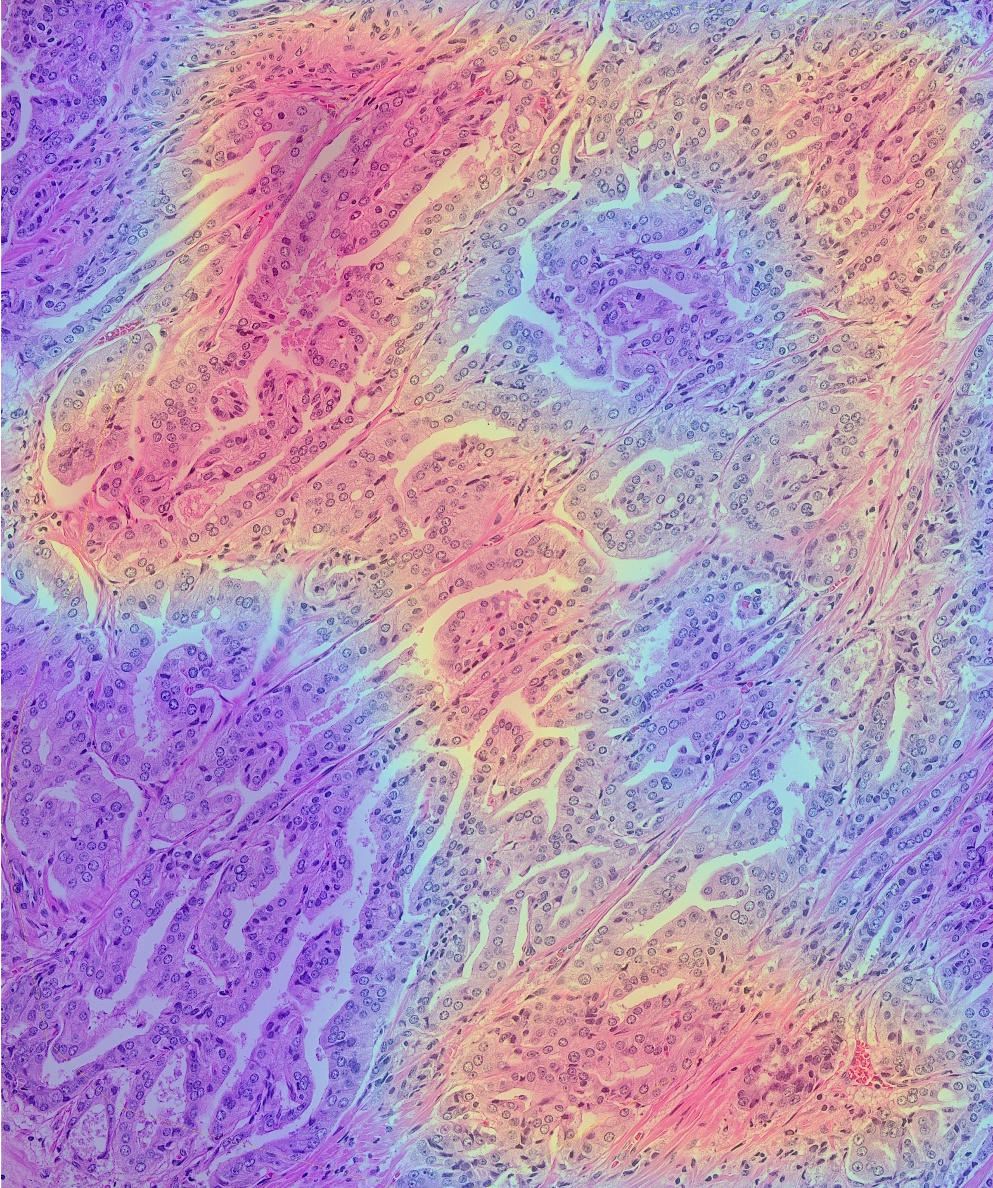} }}%
    \hfill
    \subfloat[Heatmap generated from the model optimized with $\mathcal{L}_{\text{t}}$.\label{quan:s}]{{\includegraphics[width=.2\linewidth]{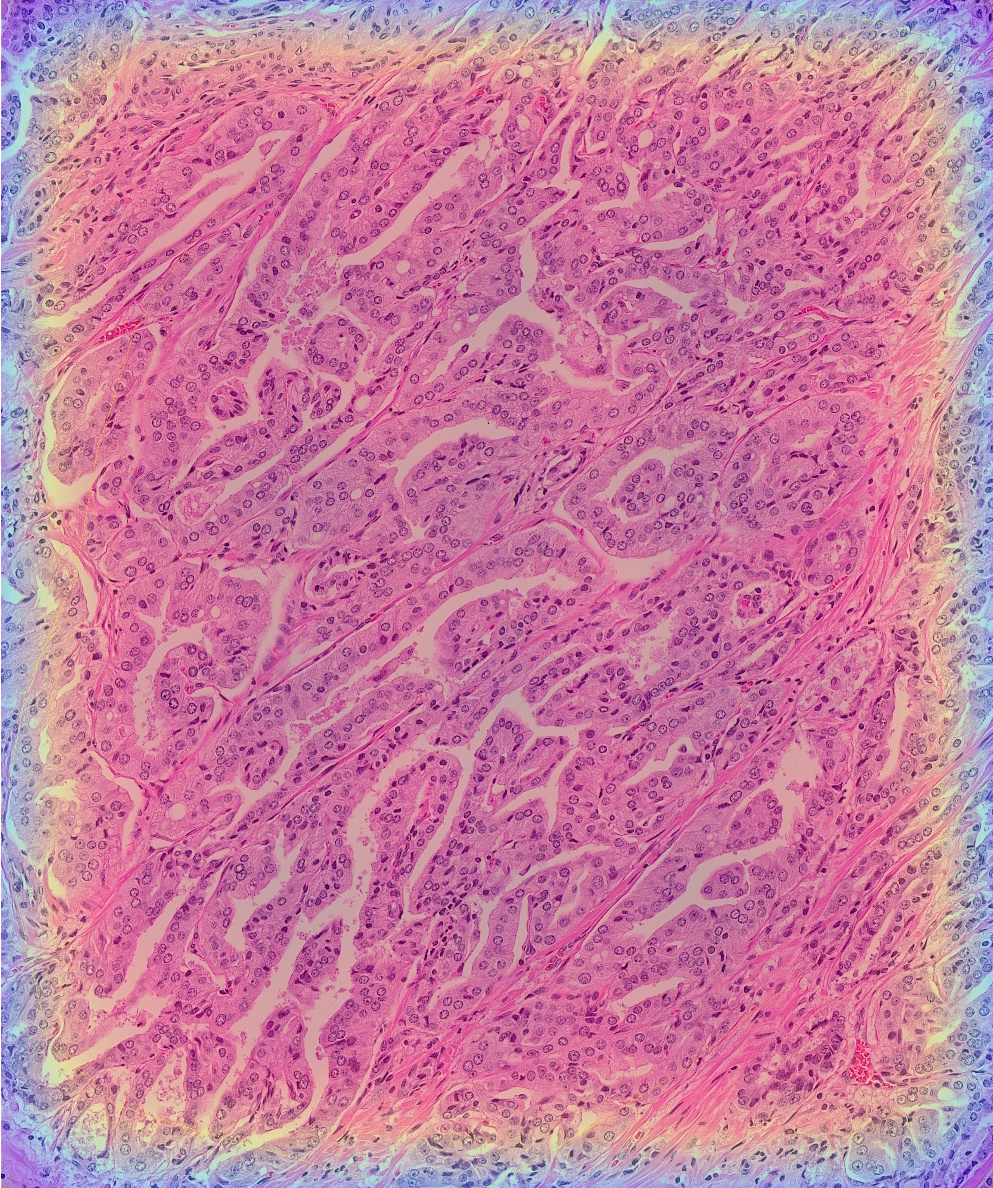} }}%
    \caption{We show an example image from CINJ with high Gleason grade and the heatmap generated from the prediction models.}%
    \label{fig:quan}%
     \vspace*{-3mm}

\end{figure}

%--------------------------------testing results-----------------------------------------
\begin{table}[tb]
    	\begin{minipage}{.45\linewidth}
      	\centering
        
        \begin{tabular}{|  C{2.2cm} | C{2.2cm} |  }
		\hline
 		& Accuracy (\%)   \\ \hline\hline
		Baseline& 54.3 \\ \hline
		$\mathcal{L}_{\text{a}}$ only & {71.4} \\ \hline
		$\mathcal{L}_{\text{t}}$ &  \textbf{77.1}\\ \hline
		\end{tabular}
        
        \caption{The unsupervised adaptation of TCGA (w/o UP) $\to$ UP. }
        \label{tcga_up}
    	\end{minipage}%
    	\hfill
    	\begin{minipage}{.45\linewidth}
      	\centering
        \begin{tabular}{|  C{2.2cm} | C{2.2cm} |  }
		\hline
 		& Accuracy (\%)   \\ \hline\hline
		Baseline& 56.3\\ \hline
		$\mathcal{L}_{\text{a}}$ only & {62.5} \\ \hline
		$\mathcal{L}_{\text{t}}$ &  \textbf{75.0} \\ \hline
		\end{tabular}
        \caption{The unsupervised adaptation of TCGA $\to$ CINJ.}
        \label{tcga_cinj}
    \end{minipage} 
     \vspace*{-6mm}
\end{table}
%--------------------------------testing results-----------------------------------------
 \vspace*{-3mm}
\paragraph{\textbf{Adaptation of TCGA $\to$ CINJ}}
The results showing in Table~\ref{tcga_cinj} also proves $\mathcal{L}_{\text{t}}$ could achieve the best adaptation performance. The confusion matrices are shown in Figure~\ref{confusion:CINJ_b}-\ref{confusion:CINJ_a}. 
We further show the qualitative results in Figure~\ref{fig:quan}. We use the probability predicted by the network on the patches to generate a Gaussian heatmap and overlay the heatmap on the original image. The red color indicates the high Gleason grade and blue color indicates the low Gleason grade. Figure~\ref{quan:o} shows an example prostate WSI from CINJ with the high Gleason grade (Gleason score 8) and the ground-truth heatmap overlaid on it. 
The heatmap generated from the baseline network is shown in Figure~\ref{quan:b}. The heatmap indicates many low Gleason grade areas, which are misclassified.
The heatmap obtained from the target network that optimized by $\mathcal{L}_{\text{a}}$ is shown in  Figure~\ref{quan:a}, which presents less low Gleason grade areas. Using $\mathcal{L}_{\text{t}}$, the target network could correctly classify all the patches into high Gleason grade, as demonstrated in Figure~\ref{quan:s}. 

% %--------------------------------confusion matrix------------------------------------------
% %--------------------------------confusion matrix------------------------------------------

\section{Conclusion}
In this work, we adopt an adversarial training and Siamese architecture to improve the classification performance of a target network in an unsupervised manner.
We show that by using the proposed domain adaptation method statistically significant classification results can be achieved. 
Future work will include improvement of the method by using extensive datasets and extension to a wide range of histopathology image classification problems. 
% %--------------------------------qualitative results------------------------------------------
% %--------------------------------qualitative results---------------------------------------------
{\small
\bibliographystyle{splncs}
\bibliography{ms.bbl}
}

\end{document}